\documentclass[letterpaper,journal]{IEEEtran}
\usepackage{amsmath,amsfonts}
\usepackage{algorithmic}
\usepackage{array}
\usepackage{textcomp}
\usepackage{stfloats}
\usepackage{url}
\usepackage{verbatim}
\usepackage{graphicx}
\usepackage{booktabs}
\usepackage{xcolor}

\newif\ifarxiv
\arxivtrue

\def\BibTeX{{\rm B\kern-.05em{\sc i\kern-.025em b}\kern-.08em
    T\kern-.1667em\lower.7ex\hbox{E}\kern-.125emX}}
\usepackage{balance}

\begin{document}
\raggedbottom

\title{Diagnosing Corruption-Induced Reliability Failures in Vision-Language Models}
\author{Xiangjie Sui, Songyang Li, Hanwei Zhu, and Baoliang Chen\\
Yuming Fang, \IEEEmembership{Fellow, IEEE}, and Xin Sun, \IEEEmembership{Member, IEEE}%
\thanks{Xiangjie Sui, Songyang Li, and Xin Sun are with the City University of Macau, Macau SAR, China.}%
\thanks{Baoliang Chen is with Nanyang Technological University, Singapore.}%
\thanks{Hanwei Zhu and Yuming Fang are with Jiangxi University of Finance and Economics, Nanchang, China.}%
\thanks{Corresponding author: Hanwei Zhu (zhuhanwei@jxufe.edu.cn).}%
}

\markboth{Submitted to IEEE Transactions on Multimedia}%
{Diagnosing Corruption-Induced Reliability Failures in Vision-Language Models}

\maketitle

\begin{abstract}
Visual corruptions can change vision--language model (VLM) behavior in ways that top-1 accuracy does not capture. A model may keep the same answer while losing distributional support, or improve accuracy through unstable wrong-to-correct changes. We introduce \textsc{Bench-C}, a controlled multiple-choice testbed for studying these effects. It selects semantically diverse samples whose predictions respond to corruption, and evaluates them under 19 corruption types and five severity levels. To measure how corruption changes the option distribution, we introduce the Robustness Alignment Score (RAS), which combines confidence--correctness alignment with uncertainty direction. We further separate originally correct samples from originally wrong samples, and track whether changes are temporary or persistent across severity. Experiments across 13 VLMs reveal a counterintuitive pattern: mild corruptions can improve top-1 accuracy while degrading prediction structure. These failures include silent degradation, erroneous overconfidence, and severity-dependent persistence. \textsc{Bench-C} therefore supports robustness evaluation that goes beyond final answers and attributes where reliability changes occur. Code and data are available at \url{https://github.com/xiangjieSui/Bench-C}.
\end{abstract}


\section{Introduction}
\IEEEPARstart{V}{ision}--language models are increasingly used to answer questions about real-world images, yet real images are rarely clean. Compression artifacts, adverse weather, sensor noise, and blur can change the visual evidence available to a model~\cite{li2024rbench, usama2025}. Humans can often still recognize and reason about a degraded scene~\cite{recht2018, hendrycks2019robustness}, but modern VLMs remain sensitive to such corruptions~\cite{li2024rbench, usama2025, xing2025demystifying}. This raises a basic question: how should we evaluate VLM reliability when the visual input is degraded?

\begin{figure*}[!t]
\centering
\includegraphics[width=0.940\textwidth]{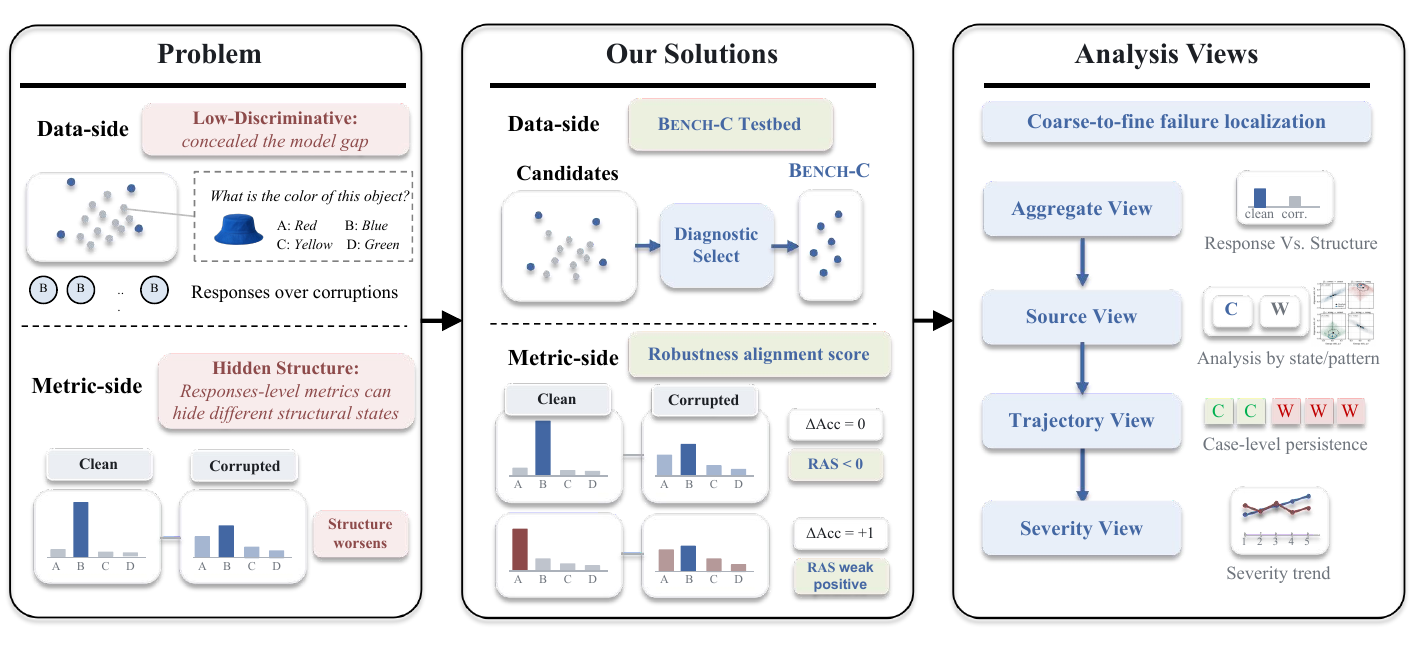}
\caption{From hidden shifts to interpretable robustness analysis. \textsc{Bench-C} focuses on corruption-sensitive samples, measures option-distribution changes, and traces robustness gaps to sample source and severity pattern.}
\label{fig:intro_overview}
\end{figure*}

Current multiple-choice robustness evaluations~\cite{li2024rbench, usama2025} provide controlled protocols, but two issues remain. First, broad benchmarks can include many \textit{low-discriminative samples}: samples whose answers stay almost unchanged across corruptions or across models. These samples may be visually simple, dominated by text priors, or insensitive to the tested degradations. They are useful for general evaluation, but they provide weak evidence for corruption robustness. The left panel of Fig.~\ref{fig:intro_overview} shows this issue with a color-identification example whose answer remains stable and therefore hides model differences. Second, top-$1$ accuracy only records the final answer. It does not show how probability is distributed across options. We refer to this option-level distribution as the \textit{prediction structure}. It indicates whether an answer is strongly supported, weakly supported, or overconfidently wrong. The middle panel of Fig.~\ref{fig:intro_overview} illustrates this gap: the final answer may stay unchanged while its support weakens, or a wrong answer may become correct without strong distributional support. Thus, the same accuracy change can reflect different reliability changes. This distinction is relevant to the visual quality paradox~\cite{xing2025demystifying}, where mild corruptions sometimes improve VLM accuracy. Accuracy alone may not tell whether such gains reflect reliable correction or unstable answer changes.

These issues motivate a focused evaluation with three goals. It should identify samples that reveal robustness differences, measure how the option distribution changes after corruption, and explain what produces the observed accuracy change. We therefore propose \textsc{Bench-C}, a controlled testbed for evaluating corruption-induced reliability failures in VLMs. \textsc{Bench-C} selects semantically diverse samples from existing multiple-choice benchmarks and keeps samples whose predictions are sensitive to corruption. We pair \textsc{Bench-C} with an analysis of clean--corrupted option distributions. We further propose the Robustness Alignment Score (RAS) to summarize whether corruption makes the option distribution more or less reliable. To turn an aggregate robustness change into its source, we analyze structural shift conditioned on whether samples were originally correct or originally wrong. To distinguish unstable changes from sustained ones, we track whether changes are absent, temporary, or persistent across severity. The right panel of Fig.~\ref{fig:intro_overview} summarizes these coarse-to-fine analysis views, from aggregate response--structure comparison to source, trajectory, and severity-level diagnosis.

We evaluate $13$ VLMs across $19$ corruption types and $5$ severity levels. The analysis shows that corruption-induced reliability changes are more structured than accuracy suggests: preserved answers can lose distributional support, and apparent accuracy gains can coexist with broader degradation on originally correct samples. In summary, our contributions are:
\begin{itemize}
\setlength{\itemsep}{0pt}
\setlength{\parsep}{0pt}
\item We build \textsc{Bench-C}, a compact corruption testbed that selects informative multiple-choice samples and applies a fixed corruption protocol.
\item We introduce RAS to measure whether corruption makes the option distribution more or less reliable, complementing response-level metrics such as accuracy and consistency.
\item We provide state-conditioned and severity-trajectory analyses that trace apparent accuracy gains to their sources and distinguish unstable changes from sustained ones.
\end{itemize}

\section{Related Work}

\subsection{MCQ Benchmarks for VLMs}

Existing multiple-choice benchmarks for VLMs cover text-rich visual comprehension and OCR~\cite{li2024seedbench2plus}, spatial reasoning~\cite{tang2025lego}, general multimodal understanding~\cite{MMBench,ying2024mmtbench}, real-world scenes~\cite{mme_realworld25}, and fine-grained visual reasoning~\cite{chen2024mmstar}. Analyses show that some questions can be answered from language priors, superficial cues, or dataset artifacts with limited image understanding~\cite{brown2025}. VQA bias studies further connect such failures to biased samples and cross-modal biases~\cite{ouyang2022suppressing,pan2024unbiased,vosoughi2024cross}, while debiased VQA and test-time adaptation studies similarly treat biased or weakly visual samples as a robustness risk~\cite{wen2024testtime}. Such low-visual-dependence samples may remain stable under degradation and fail to reveal robustness gaps. \textsc{Bench-C} keeps the multiple-choice format but uses it for paired clean-to-corrupted comparison over a fixed answer set. Rather than estimating general capability, it selects samples that expose model changes under visual corruption.

\subsection{Corruption Robustness}
Corruption robustness has long been studied in computer vision. \textsc{ImageNet-C}~\cite{hendrycks2019robustness} showed that high clean accuracy does not prevent degradation under common corruptions. Recent VLM studies evaluate real-world or common corruptions~\cite{li2024rbench,qiu2025imagecorruptions,usama2025}, broaden perturbation and severity coverage~\cite{saxena2026vlmrobustbench}, and test visual-quality judgment~\cite{zhu2024twafc}. The visual quality paradox further shows that moderate degradation can sometimes improve VLM accuracy~\cite{xing2025demystifying}. These studies establish broad corruption coverage and accuracy-level failure patterns. We also use controlled corruptions, but ask whether an observed response change is supported by the option distribution and which samples produce it.

\subsection{Compact and Diagnostic Benchmarking}
Compact benchmarking reduces evaluation cost by selecting informative samples. For example, tinyBenchmarks~\cite{polo2024tinybenchmarks} uses item response theory to choose subsets that approximate larger evaluation sets. \textsc{Bench-C} uses a related principle, but selects samples that reveal behavior changes under visual corruption rather than samples that approximate aggregate capability. Diagnostic evaluation provides a complementary view. VQA models can rely on dense inter- and intra-modality interactions and adaptive reasoning modules, so final correctness alone need not reveal the underlying answer process~\cite{liu2021dense,zhong2021selfadaptive}. Calibration~\cite{guo2017calibration}, proper scoring rules~\cite{gneiting2007proper}, and likelihood-shift analyses~\cite{oren2023contamination,shi2023detecting} show that distribution-level signals can reveal behaviors hidden from accuracy.

These two directions leave a gap for VLM corruption robustness. Compact benchmarks usually estimate general capability with fewer samples, while distribution-level diagnostics are usually not built around matched clean--corrupted visual inputs. \textsc{Bench-C} addresses this gap by selecting corruption-responsive samples and measuring how correctness, uncertainty, and confidence--correctness alignment change under controlled degradation.

\subsection{Complementary Robustness Paradigms}
Robustness evaluation also studies threat models beyond common visual corruptions, including adversarial traceability~\cite{fang2023tracing}, transferable attacks~\cite{li2023sibling}, defenses based on augmentation, pruning, or visual representation masking~\cite{frosio2023best,Piras_2025,liu2025improving}, answer-option perturbations~\cite{zong2023permutations}, and multimodal tampering~\cite{agarwal2024mvtamperbench}. These settings may change the perturbation strategy, text input, or cross-modal consistency. In contrast, our setting changes only the image while keeping the question and answer options fixed, making the measured shift easier to attribute to visual degradation.

\section{Diagnostic Problem Formulation}
Let $q=(I,x,\mathcal{O},y)$ denote a visual multiple-choice question, where $I$ is the image, $x$ is the textual question, $\mathcal{O}=\{o_k\}_{k=1}^{K}$ is the fixed option set, and $y\in\{1,\ldots,K\}$ is the ground-truth option index. Given $q$, a VLM produces an option-level score vector, which is normalized into a distribution $p\in\mathbb{R}^{K}$ over the $K$ candidate answers. The predicted answer is $\hat{y}=\arg\max_k p_k$.

For a corruption type $d\in\mathcal{D}$ and severity level $\ell\in\{1,\ldots,5\}$, let $T_{d,\ell}(I)$ denote the corrupted image obtained by applying corruption $d$ at severity $\ell$ to the clean image $I$. We denote the clean prediction distribution as $p^{(o)}$ and the corrupted prediction distribution as $p^{(d,\ell)}$. This bounded option space enables explicit clean--corrupted comparisons over matched answer distributions. We treat the normalized option scores as diagnostic distributions, not fully calibrated posteriors, and all conclusions are conditioned on fixed option content, fixed option order, and unchanged textual prompts. This scope makes the source of the measured change identifiable: only the visual input is perturbed. Under this setting, corruption robustness is not only a question of whether the final answer changes. Instead, it involves three key questions:
\begin{itemize}
\item \textit{{Which samples reveal robustness gaps?}} Identifying such samples makes the testbed sensitive to corruption-induced model differences, rather than dominated by examples whose answers barely change.

\item \textit{{How does prediction structure change under corruption?}} Measuring option-level shifts helps reveal prediction-structure changes that can be hidden by unchanged answers or aggregate accuracy.

\item \textit{{What produces the observed accuracy change?}} Attributing the shift helps localize where robustness changes occur, understand model behavior under corruption, and guide future training.
\end{itemize}
These questions define the three parts of our analysis. The testbed construction selects samples that are informative under corruption. The prediction-structure metrics measure how the clean and corrupted option distributions differ. The coarse-to-fine analysis then traces aggregate shifts to their source, state, and severity pattern.

\section{\textsc{Bench-C} Testbed Construction}
\begin{figure*}[t]
    \centering
\includegraphics[width=1\linewidth]{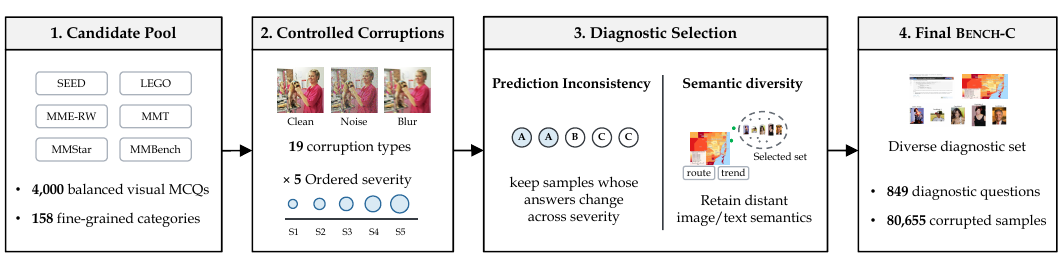}
\caption{Construction pipeline of the \textsc{Bench-C} testbed. Six source multiple-choice VLM benchmarks are balanced into a candidate pool and evaluated under $19$ corruptions and five severity levels. Diagnostic selection keeps corruption-responsive and semantically diverse samples, yielding $849$ questions and $80{,}655$ corrupted samples.}    \label{fig:bench_c}
\end{figure*}

\textsc{Bench-C} is a controlled testbed for analyzing how visual degradation changes VLM decisions. Its construction has three goals: keep semantically diverse samples, apply controlled corruptions with ordered severity levels, and select samples that respond to corruption. Fig.~\ref{fig:bench_c} summarizes the pipeline: candidate samples are corrupted, scored by answer variation across severity, and retained when they also improve image and text semantic coverage.

\subsection{Candidate Pool}
We construct the candidate pool from six representative multiple-choice VLM benchmarks: \textsc{SeedBench2+}~\cite{li2024seedbench2plus}, \textsc{LEGO-Puzzles}~\cite{tang2025lego}, \textsc{MME-RealWorld-Lite}~\cite{mme_realworld25}, \textsc{MMT-Bench-VA}~\cite{ying2024mmtbench}, \textsc{MMStars}~\cite{chen2024mmstar}, and \textsc{MMBench-DEV-EN-V11}~\cite{MMBench}. These sources cover text-rich, spatial, real-world, commonsense, OCR, and fine-grained visual reasoning. We balance the pool across source benchmarks and task families, yielding an initial set $\mathcal{A}$ of $4{,}000$ samples.

\subsection{Controlled Corruption Protocol}

\textsc{Bench-C} applies one visual corruption at a time over five ordered severity levels, following the common-corruption convention used in \textsc{ImageNet-C}~\cite{hendrycks2019robustness}. The protocol contains $19$ corruption types grouped into five families: photometric changes, blur, noise, weather, and digital artifacts. For a candidate question $q=(I,x,\mathcal{O},y)$, each corrupted instance replaces only the image $I$ with $T_{d,\ell}(I)$ while keeping the question $x$ and option set $\mathcal{O}$ fixed. Using one corruption at a time makes the source of change easier to interpret. Mixed or naturally occurring degradations may better approximate some deployment scenarios, but they can entangle multiple visual factors. The controlled protocol supports family- and severity-specific analysis of model behavior.

\subsection{Corruption-Induced Prediction Inconsistency}

For each candidate $q$, selector model $f_s\in\mathcal F$, and corruption type $d$, we collect top-$1$ predictions over the five severity levels,
$\hat{\mathbf y}^{(d)}_{f_s}(q)
=
\left\{
\hat y^{(d,1)}_{f_s}(q),\ldots,
\hat y^{(d,5)}_{f_s}(q)
\right\}.$
Let $\pi_{k,f_s}^{(d)}(q)$ be the frequency with which option $k$ appears in this trajectory. We measure trajectory dispersion with normalized Gini impurity:
\begin{equation}
G_{f_s}^{(d)}(q)
=
\frac{K}{K-1}
\left[
1-\sum_{k=1}^{K}\left(\pi_{k,f_s}^{(d)}(q)\right)^2
\right].
\end{equation}
We average this dispersion over corruption types and selector models to obtain the corruption-responsiveness score:
\begin{equation}
\kappa(q)=
\frac{1}{|\mathcal{D}|\,|\mathcal{F}|}
\sum_{d\in\mathcal{D}}\sum_{f_s\in\mathcal{F}}
G_{f_s}^{(d)}(q).
\end{equation}
Higher $\kappa(q)$ means that corruptions produce more varied top-$1$ answers across selectors and severity levels. We use MoonDream2~\cite{moondream2025}, InternVL3-1B~\cite{Internvl2024l}, and Ovis2-1B~\cite{2024ovis} as selectors. Gini impurity is suitable here because selection only needs categorical answer variation. Importantly, $\kappa$ measures responsiveness to corruption, not intrinsic difficulty: a sample answered incorrectly at every severity has zero trajectory dispersion, whereas an initially easy sample can score highly if corruption destabilizes its answer.

\subsection{Semantic Novelty and Final Selection}

We first keep candidates for which every selector shows nonzero average severity dispersion,
\begin{equation}
\mathcal A^{+}
=
\left\{
q\in\mathcal A:
\frac{1}{|\mathcal D|}
\sum_{d\in\mathcal D}
G_{f_s}^{(d)}(q)>0,
\ \forall f_s\in\mathcal F
\right\}.
\end{equation}
This response filter leaves $1{,}750$ candidates. Requiring evidence from every selector is stricter than thresholding only the pooled average $\kappa$; it prevents one selector's large response from hiding zero response in another selector. Responsiveness alone can over-concentrate the selected set on one fragile task type or a cluster of near-duplicates. We therefore add a semantic diversity term. We compute image and question embeddings using pretrained visual~\cite{dosovitskiy2020vit} and sentence encoders~\cite{reimers2019sentencebert}, and perform greedy semantic selection. Let $v_q^I$ and $v_q^T$ denote the image and text embeddings of $q$, and let $\mu_{n-1}^I$ and $\mu_{n-1}^T$ be the centroids of the already selected set $\mathcal A_{n-1}^{\star}$. The semantic novelty of a candidate is:
\begin{equation}
\nu_n(q)
=
1-
\frac{
\cos(v_q^I,\mu_{n-1}^I)+\cos(v_q^T,\mu_{n-1}^T)
}{2},
\end{equation}
where both similarities are initialized to zero for the first selection step. We traverse the eligible pool $\mathcal A^{+}$ and assign each item the novelty value observed when it is selected relative to the preceding semantic coverage: 
\begin{equation}
q_n^\star
=
\arg\max_{q\in\mathcal A^{+}\setminus\mathcal A_{n-1}^{\star}}
\nu_n(q).
\end{equation}
The final selection score combines the behavioral term and the semantic term without tuning downstream model performance: $e(q)=\kappa(q)+\nu(q)$. We select candidates above the mean score in the eligible pool:
\begin{equation}
\mathcal Q_{\textsc{Bench-C}}
=
\left\{
q\in\mathcal A^{+}:
e(q)>\mathbb E_{q'\in\mathcal A^{+}}[e(q')]
\right\}.
\end{equation}
This yields $849$ samples ($21.2\%$ of the candidate pool). Applying $19$ corruptions at five severity levels produces $849\times19\times5=80{,}655$ corrupted image--question pairs, in addition to the clean condition.

\subsection{Validation Analysis}
\begin{figure}[t]
    \centering

\includegraphics[width=0.72\columnwidth]
{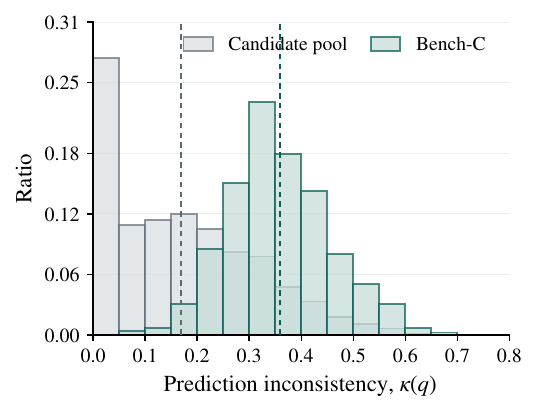}
\par\vspace{1pt}
{\footnotesize (a) Distribution of corruption-induced prediction inconsistency.}

\vspace{2pt}

\includegraphics[width=0.8\columnwidth]
{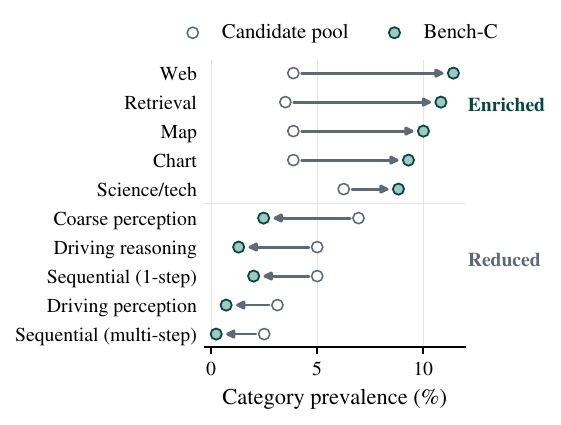}
\par\vspace{1pt}
{\footnotesize (b) Category-distribution shifts from the candidate pool to \textsc{Bench-C}.}

    \caption{Selection diagnostics for \textsc{Bench-C}.
    The selected testbed is enriched for corruption-induced prediction
    inconsistency while intentionally shifting toward visually detailed and structured task categories.}
    \label{fig:selection-diagnostics}
\end{figure}

The validation analysis checks whether the selected set has the intended properties. Fig.~\ref{fig:selection-diagnostics}(a) shows that \textsc{Bench-C} shifts away from the large low-inconsistency mass in the $4{,}000$-question pool, indicating that selection enriches corruption-responsive samples. Fig.~\ref{fig:selection-diagnostics}(b) shows that selection also changes the task mix: visually detailed and structured tasks, including web, retrieval, map, and chart understanding, become more common, while coarse perception and driving-related categories become less common. This confirms that \textsc{Bench-C} is intentionally reweighted toward samples that are sensitive to visual corruption.
To test whether the selected samples are dominated by a single selector, we conduct a leave-one-selector-out test. Table~\ref{tab:selection-validity} shows that most released questions are retained when the original eligibility pool is fixed (JI $>0.710$), and a majority remains after recomputing eligibility (JI $>0.606$). We also compare semantic-only and inconsistency-only selection, which overlap at a JI of $0.646$, indicating that the two criteria select related but distinct samples. These checks indicate that the selected set is not driven by a single selector or selection signal. See Supplementary Material, Sec.~II, for composition and $\kappa(q)$ examples.


\section{Prediction-Structure Diagnostics}


The second question asks how the option-level prediction structure changes when the image is degraded. Accuracy only records whether the top-$1$ answer is correct, but the same accuracy transition can have different reliability meanings: a correct answer may become weakly supported, a wrong answer may become overconfident, or an originally wrong answer may be corrected with only fragile support. These cases are indistinguishable under accuracy alone. We therefore compare the clean and corrupted option distributions for the same question. This comparison lets us measure not only whether the answer changes, but also whether the distribution supporting that answer becomes more or less reliable.

\begin{table}[t]
\centering
\caption{Leave-one-selector-out sensitivity of \textsc{Bench-C}. Jaccard index (JI) is measured against the released split; fixed variants retain the original $1{,}750$-candidate pool, whereas recomputed variants update eligibility after removing each selector.}
\label{tab:selection-validity}
\footnotesize
\setlength{\tabcolsep}{5pt}
\begin{tabular}{lcccc}
\toprule
Removed
& Fixed-JI
& Fixed-Ret.
& Recomp.-JI
& Recomp.-Ret. \\
\midrule
InternVL3-1B
& $0.747$
& $726/849$
& $0.685$
& $690/849$ \\

Moondream2
& $0.710$
& $705/849$
& $0.606$
& $641/849$ \\

Ovis2-1B
& $0.749$
& $727/849$
& $0.658$
& $674/849$ \\
\bottomrule
\end{tabular}
\end{table}

\subsection{Matched Option Distributions}

For each question $q=(I,x,\mathcal O,y)$, the clean image $I$ and the corrupted image $T_{d,\ell}(I)$ share the same textual question $x$ and option set $\mathcal O$. A VLM produces two comparable option distributions: the clean distribution $p^{(o)}$ and the corrupted distribution $p^{(d,\ell)}$. Raw option scores are converted into a valid distribution over the $K$ answer choices by applying softmax to logits. The same conversion is used for the clean and corrupted versions of a question. We use these distributions for paired within-model comparison, not as calibrated posteriors comparable across model families.


\subsection{Distribution-Level Components}

We characterize prediction-structure change with two components: uncertainty shift and confidence--correctness alignment shift. Uncertainty shift describes whether the corrupted prediction becomes flatter or sharper. Alignment shift describes whether the model's confidence becomes better or worse matched to correctness. Together, these components distinguish cases that have the same top-$1$ outcome but different reliability meanings.

\begin{figure*}[t]
    \centering
    \includegraphics[width=0.92\linewidth]{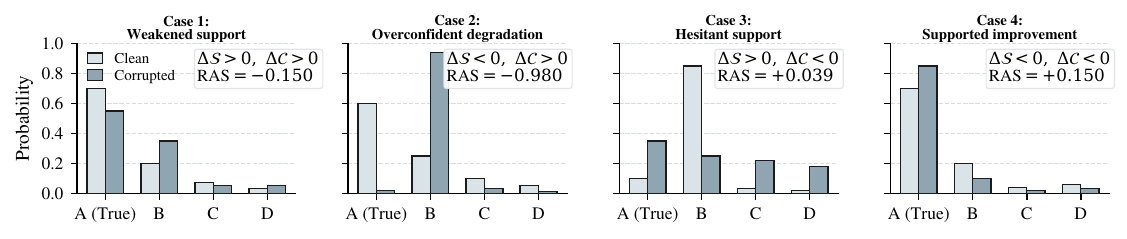}
    \vspace{-2pt}
\caption{Representative prediction-structure shifts under corruption. Each panel compares clean and corrupted option distributions, with option A as ground truth. The four cases correspond to $\Delta\mathcal S$ and $\Delta\mathcal C$ sign patterns: weakened support, overconfident degradation, hesitant support, and supported improvement.}
    \label{fig:ras}
\end{figure*}

\textit{(1) Uncertainty shift} $\Delta\mathcal S$.
To measure how decisive an option distribution is, we use normalized entropy:
\begin{equation}
\mathcal S(p)
=
\frac{-\sum_{k=1}^{K}p_k\log p_k}{\log K},
\end{equation}
where $\mathcal S(p)\in[0,1]$. A lower value indicates that probability mass is concentrated on fewer options, while a higher value indicates a flatter distribution over the option set. The uncertainty shift is defined as:
\begin{equation}
\Delta\mathcal S^{(d,\ell)}
=
\mathcal S(p^{(d,\ell)})-\mathcal S(p^{(o)}).
\end{equation}
Thus, $\Delta\mathcal S>0$ means that corruption makes the prediction flatter and less decisive, whereas $\Delta\mathcal S<0$ means that corruption sharpens the prediction.

\textit{(2) Confidence--correctness alignment shift} $\Delta\mathcal{C}$.
Uncertainty direction alone is not enough for interpreting reliability change. Increased uncertainty may reflect either appropriate caution or loss of discrimination; decreased uncertainty may reflect either stronger evidence or overconfident error. We therefore interpret it together with confidence--correctness alignment shift.
For an option distribution $p$ and ground-truth label $y$, the shift is
\begin{equation}
\mathcal C(p,y)
=
\left|
\mathbf{1}\{\arg\max_k p_k=y\}
-
\max_k p_k
\right|.
\end{equation}
This quantity compares the model's top-$1$ confidence with the correctness of its top-$1$ prediction. If the prediction is correct, the shift is small when the model assigns high confidence to the predicted answer. If the prediction is wrong, the shift is small when the model avoids assigning high confidence to the wrong answer. The confidence--correctness alignment shift induced by corruption is defined as:
\begin{equation}
\Delta\mathcal C^{(d,\ell)}
=
\mathcal C(p^{(d,\ell)},y)
-
\mathcal C(p^{(o)},y).
\end{equation}
A positive $\Delta\mathcal C$ indicates degraded alignment under corruption, whereas a negative value indicates improved alignment. Here $\mathcal C$ is a sample-level residual for matched clean--corrupted comparison, not a population calibration estimator such as expected calibration error.

\subsection{Robustness Alignment Score}
We define RAS as a paired summary of reliability change. The main term is the negative alignment shift, $-\Delta\mathcal C$: improved confidence--correctness alignment increases RAS, while degraded alignment decreases it. The uncertainty term changes the interpretation of two cases. If alignment worsens while entropy decreases, the model becomes more confidently wrong and receives an additional penalty. If alignment improves while entropy increases, the improvement is treated as weaker support.
Formally, RAS for a clean--corrupted pair is defined as (we omit the superscript $(d,\ell)$ for simplicity):
\begin{equation}
\mathrm{RAS}
= -\Delta \mathcal{C} 
- \underbrace{[\Delta \mathcal{C}]_{+}[-\Delta \mathcal{S}]_{+}}_{\text{Overconfidence penalty}}
- \underbrace{[\Delta \mathcal{S}]_{+} [-\Delta \mathcal{C}]_{+}}_{\text{Hesitation penalty}},
\end{equation}
where $[x]_{+}=\max(0,x)$. Since both $\Delta\mathcal S$ and $\Delta\mathcal C$ lie in $[-1,1]$, RAS lies in $[-2,1]$. Higher RAS values indicate reliable prediction-structure behavior under corruption. A positive RAS indicates improved alignment after the uncertainty adjustment, whereas a negative RAS indicates degraded alignment or an uncertainty penalty. The asymmetric range reflects the risk-sensitive design of RAS: positive values come only from improved alignment, whereas negative values can include penalties for overconfident degradation or hesitant support. RAS can separate clean--corrupted shifts that have the same top-$1$ outcome but different reliability meanings. Fig.~\ref{fig:ras} gives representative examples; Supplementary Material, Sec.~III-A, gives sign-regime and case-level checks.


\section{Attributing Structural Change}

The third question asks where the change comes from. Accuracy reports an aggregate change, which can mix damage to originally correct samples, correction of originally wrong samples, and hidden structural shifts when top-$1$ correctness does not change. We therefore attribute changes to their clean-side source and refine each source into correctness states. This view measures structural change within each state and provides the basis for the severity-trajectory analysis in experiments.

\subsection{Source Attribution}
We use the clean prediction to define two sources of change: originally correct samples $\mathcal Q_{c}^{(d,\ell)}=\{q:\hat y^{(o)}=y\}$ and originally wrong samples $\mathcal Q_{w}^{(d,\ell)}=\{q:\hat y^{(o)}\neq y\}$. This attribution separates damage to initially reliable behavior from correction of initially wrong behavior, instead of merging both into one aggregate response shift.

\subsection{State Attribution}
Within each source, we further group samples by whether the corrupted prediction is correct. We use two-letter subscripts: the first letter denotes clean correctness, and the second denotes corrupted correctness. Here, $c$ denotes a correct top-$1$ prediction and $w$ denotes an incorrect one:
\begin{equation}
\label{eq:correctness-state-partition}
\begin{aligned}
\mathcal Q_{cc}^{(d,\ell)}
&=\{q:\hat y^{(o)}=y,\ \hat y^{(d,\ell)}=y\},\\
\mathcal Q_{cw}^{(d,\ell)}
&=\{q:\hat y^{(o)}=y,\ \hat y^{(d,\ell)}\neq y\},\\
\mathcal Q_{wc}^{(d,\ell)}
&=\{q:\hat y^{(o)}\neq y,\ \hat y^{(d,\ell)}=y\},\\
\mathcal Q_{ww}^{(d,\ell)}
&=\{q:\hat y^{(o)}\neq y,\ \hat y^{(d,\ell)}\neq y\}.
\end{aligned}
\end{equation}
The four states separate visible answer changes from hidden structural changes. The $\mathcal Q_{cw}$ and $\mathcal Q_{wc}$ groups capture top-$1$ flips: the former contains failures caused by corruption, and the latter contains corrections. The $\mathcal Q_{cc}$ and $\mathcal Q_{ww}$ groups keep the same top-$1$ correctness state under corruption, revealing whether corruption weakens a still-correct prediction or improves a still-wrong one.

\subsection{State-Conditioned Structural Effects}

We report structural effects at three levels: aggregate, source, and state. The aggregate effect gives the overall direction under a corruption condition. The source effects split this direction by originally correct and originally wrong samples. The state effects further split each source by the corrupted correctness state. These quantities are computed from state-conditioned averages.
For any diagnostic quantity $M\in\{\mathrm{RAS},\Delta\mathcal S,\Delta\mathcal C\}$, the state-conditioned average is
\begin{equation}
\overline{M}_{ab}^{(d,\ell)}
=
\mathbb E_{q\in\mathcal Q_{ab}^{(d,\ell)}}
\left[
M^{(d,\ell)}(q)
\right],
\qquad a,b\in\{c,w\}.
\end{equation}
The within-source and aggregate weights are
\begin{equation}
\begin{aligned}
\alpha_{ab}^{(d,\ell)}
&=
\frac{|\mathcal Q_{ab}^{(d,\ell)}|}
{|\mathcal Q_a^{(d,\ell)}|},
\qquad
\pi_a^{(d,\ell)}
=
\frac{|\mathcal Q_a^{(d,\ell)}|}{|\mathcal Q|}.
\end{aligned}
\end{equation}
Using $\alpha_{ab}$, the source-level effect averages the states within the same clean-side source:
\begin{equation}
\label{eq:side-level-structural-effect}
\overline{M}_{a}^{(d,\ell)}
=
\sum_{b\in\{c,w\}}
\alpha_{ab}^{(d,\ell)}
\overline{M}_{ab}^{(d,\ell)}.
\end{equation}
We then combine the two sources by their prevalence to obtain the aggregate effect:
\begin{equation}
\overline{M}^{(d,\ell)}
=
\sum_{a\in\{c,w\}}
\pi_a^{(d,\ell)}
\overline{M}_{a}^{(d,\ell)}.
\end{equation}
When $M=\mathrm{RAS}$, we write the source-level and aggregate summaries as $R_a$ and $\overline{\mathrm{RAS}}$. The source attribution identifies whether structural change comes from originally correct or originally wrong samples. State attribution then separates visible failures and recoveries from hidden reliability shifts. Combining the two attributions separates aggregate response change into source- and state-level structural effects.
\begin{table*}[t]
\centering
\caption{Model-level robustness diagnosis on \textsc{Bench-C}. $\mathrm{Acc}^{(o)}$ measures clean capability; $\Delta\mathrm{Acc}$, consistency, and relative robustness measure response change; RAS measures structural robustness; $(R_c,R_w)$ localize clean-correct and clean-wrong effects. Mixed is the percentage of cells with $R_c<0$ and $R_w>0$. Bold marks the column extreme.}
\label{tab:model-robustness-profile}
\footnotesize
\setlength{\tabcolsep}{2.0pt}
\renewcommand{\arraystretch}{1.05}
\begin{tabular*}{\textwidth}{@{\extracolsep{\fill}}>{\raggedright\arraybackslash}p{0.22\textwidth}cccccccc@{}}
\toprule
 & Capability & \multicolumn{3}{c}{Response} & \multicolumn{4}{c}{Structural diagnosis} \\
\cmidrule(lr){2-2}\cmidrule(lr){3-5}\cmidrule(lr){6-9}
Model & $\mathrm{Acc}^{(o)}$ & $\Delta\mathrm{Acc}$ & Consistency & Rel. Rob. & RAS & $R_c$ & $R_w$ & Mixed (\%) \\
\midrule
Emu3-Chat~\cite{wang2024emu3} & $0.352$ & $-0.041$ & $0.796$ & $0.884$ & $-0.057$ & $-0.170$ & $0.005$ & $76.8$ \\
Monkey-Chat~\cite{li2024monkey} & $0.417$ & $-0.052$ & $0.739$ & $0.875$ & $-0.068$ & $-0.167$ & $0.003$ & $51.6$ \\
Gemma3~\cite{gemma3_2025} & $0.428$ & $-0.061$ & $0.754$ & $0.858$ & $-0.073$ & $-0.234$ & $0.047$ & $100.0$ \\
Falcon2-VLM~\cite{malartic2024falcon2} & $0.441$ & $\mathbf{-0.040}$ & $0.737$ & $\mathbf{0.909}$ & $-0.058$ & $\mathbf{-0.115}$ & $-0.014$ & $\mathbf{4.2}$ \\
Eagle-X5~\cite{shi2025eagle} & $0.451$ & $-0.069$ & $0.743$ & $0.848$ & $-0.075$ & $-0.183$ & $0.014$ & $75.8$ \\
mPLUG-Owl3~\cite{ye2025mplugowl} & $0.465$ & $-0.049$ & $\mathbf{0.800}$ & $0.894$ & $\mathbf{-0.049}$ & $-0.131$ & $0.023$ & $93.7$ \\
Idefics3-Llama3~\cite{laurencon2024building} & $0.561$ & $-0.094$ & $0.707$ & $0.832$ & $-0.101$ & $-0.243$ & $\mathbf{0.080}$ & $100.0$ \\
Molmo~\cite{Deitke2025Molmo} & $0.565$ & $-0.072$ & $0.788$ & $0.873$ & $-0.079$ & $-0.171$ & $0.040$ & $100.0$ \\
SAIL-VL-1.6~\cite{dong2025sail} & $0.580$ & $-0.063$ & $0.757$ & $0.892$ & $-0.073$ & $-0.144$ & $0.024$ & $93.7$ \\
DeepSeek-VL2-Small~\cite{2024deepseekvl2} & $0.594$ & $-0.089$ & $0.798$ & $0.850$ & $-0.088$ & $-0.171$ & $0.032$ & $98.9$ \\
Phi-4-Multimodal-Instruct~\cite{abouelenin2025phi4} & $0.604$ & $-0.098$ & $0.759$ & $0.837$ & $-0.101$ & $-0.199$ & $0.050$ & $100.0$ \\
Kimi-VL-A3B-Instruct~\cite{Kteam2025kimi} & $0.630$ & $-0.096$ & $0.778$ & $0.848$ & $-0.098$ & $-0.181$ & $0.044$ & $100.0$ \\
Qwen2.5-VL-Instruct~\cite{qwen25vl_2025} & $\mathbf{0.667}$ & $-0.115$ & $0.753$ & $0.828$ & $-0.118$ & $-0.199$ & $0.045$ & $98.9$ \\
\bottomrule
\end{tabular*}
\end{table*}

\section{Experiments and Analysis}

\newenvironment{findingbox}[2]{%
    \vspace{0.1em}
    \noindent\begingroup
    \setlength{\fboxsep}{0pt}%
    \noindent\hspace{0.15em}%
    \vrule width 1.1pt\hspace{0.45em}%
    \begin{minipage}[t]{0.965\columnwidth}
    \textbf{Observation #1. #2}\quad\ignorespaces
}{%
    \par
    \end{minipage}
    \endgroup
    \vspace{0.1em}
}

\subsection{Experimental Setup}
\label{sec:experimental-setup}

We use \textsc{Bench-C} to study how visual corruption changes VLM prediction structure. The testbed contains $849$ multiple-choice questions. Each clean image is paired with $19$ single-factor corruptions at five severity levels, producing $80{,}655$ corrupted image--question pairs. Because \textsc{Bench-C} is enriched for corruption-responsive samples, the experiments focus on diagnostic patterns rather than failure prevalence in the original source benchmarks.

We evaluate $13$ VLMs with available option-level logits (see Table~\ref{tab:model-robustness-profile}). For each model, logits are softmaxed within the valid option set to obtain matched clean and corrupted option distributions. These distributions are used only for paired within-model comparisons, not as calibrated probabilities comparable across model families. For black-box models without exposed logits, our released code also supports a sampling-based estimate of the option distribution; see Supplementary Material, Sec.~III-B.

For each clean--corrupted pair, we compute clean accuracy $\mathrm{Acc}^{(o)}$, corrupted accuracy $\mathrm{Acc}^{(d,\ell)}$, accuracy shift $\Delta\mathrm{Acc}=\mathrm{Acc}^{(d,\ell)}-\mathrm{Acc}^{(o)}$, top-$1$ consistency $\mathbf{1}\{\hat y^{(d,\ell)}=\hat y^{(o)}\}$, relative robustness $\mathrm{Acc}^{(d,\ell)}/\mathrm{Acc}^{(o)}$~\cite{li2024rbench}, RAS, and the RAS components $\Delta\mathcal S$ and $\Delta\mathcal C$. Model-, corruption-, and severity-level quantities are ordinary arithmetic means. 

\begin{table}[!thp]
\caption{Four-state structural diagnosis under sample averaging. $\Delta\mathrm{Acc}$ records the visible top-$1$ transition; RAS and its components measure prediction-structure change. Share is measured over all clean--corrupted pairs, and conditional rate within the clean-correct or clean-wrong opportunity set.}
\label{tab:state-overall}
\small
\setlength{\tabcolsep}{4pt}
\begin{tabular}{@{}lrrrrrr@{}}
\toprule
State & Share & Cond. rate & $\Delta\mathrm{Acc}$ & RAS & $\Delta\mathcal S$ & $\Delta\mathcal C$ \\
\midrule
$Q_{cc}$ & $39.91\%$ & $76.82\%$ & $0$ & $-0.019$ & $0.029$ & $0.019$ \\
$Q_{cw}$ & $12.04\%$ & $23.18\%$ & $-1$ & $\mathbf{-0.709}$ & $0.106$ & $\mathbf{0.679}$  \\
$Q_{wc}$ & $4.82\%$ & $10.03\%$ & $+1$ & $\mathbf{0.462}$ & $0.012$ & $\mathbf{-0.516}$  \\
$Q_{ww}$ & $43.23\%$ & $89.97\%$ & $0$ & $-0.021$ & $0.012$ & $-0.006$  \\
\bottomrule
\end{tabular}
\end{table}

The analysis follows a coarse-to-fine route. We first compare aggregate response robustness with structural robustness. We then attribute the mismatch to source and state groups, inspect the component directions, and trace whether these effects persist across severity. Finally, we examine condition- and model-level patterns to identify which corruptions and model families drive the main effects.

\subsection{Aggregate View: Outcome and Structure Disagree}
\label{sec:overall-decoupling}

Table~\ref{tab:model-robustness-profile} starts from the aggregate view. The table shows that clean capability, response-level robustness, and structural-level robustness are not the same signal. For example, Qwen2.5-VL has the highest clean accuracy, but also the largest accuracy loss and the most negative RAS. Conversely, Falcon2 has the smallest accuracy drop and strongest relative robustness, but mPLUG-Owl3 has the best RAS and top-$1$ consistency.  These aggregate mismatches motivate the source- and state-level attribution; Supplementary Material, Secs.~IV-A--IV-C, expands them with rank-shift, component, and trajectory analyses.

\begin{findingbox}{$1$}{Answer stability can hide structural instability.}
Similar response robustness can correspond to different RAS profiles, so preserved top-$1$ answers should be checked at the distribution level.
\end{findingbox}

\subsection{Source View: Attribution Reveals the Tradeoff}
\label{sec:state-conditioned-diagnosis}

To explain the aggregate mismatch, we move from model-level averages to source and state attribution. Using the correctness states in Eq.~\eqref{eq:correctness-state-partition}, Table~\ref{tab:state-overall} reports how each state contributes to response change and structural change. This view asks whether robustness loss comes from damage to originally correct samples, and whether apparent gains reflect reliable correction of originally wrong samples.

Table~\ref{tab:state-overall} shows that the mismatch is mainly a source-attribution problem. On originally correct samples, answer-preserving cases are not neutral: many remain correct but lose structural support, while clean-to-wrong transitions carry the strongest damage. On originally wrong samples, wrong-to-correct transitions are structurally constructive, but they are much smaller than persistent wrong cases. Thus, apparent correction is limited and must be read against the broader clean-correct cost.

\begin{figure}[t]
    \centering
    \includegraphics[width=0.95\columnwidth]{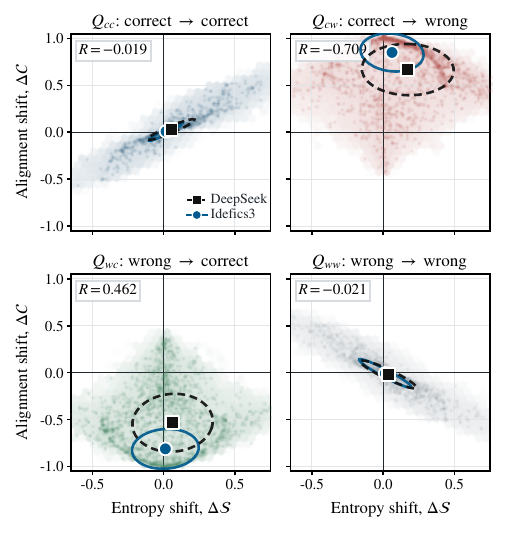}
\caption{RAS component plane by correctness state. Each panel shows clean--corrupted pairs over density contours in the $(\Delta\mathcal S,\Delta\mathcal C)$ plane; boxes report mean RAS. Blue circles and black squares mark Idefics3 and DeepSeek, whose similar accuracy loss hides different component profiles.}
    \label{fig:ras-plane-by-state}
\end{figure}

Fig.~\ref{fig:ras-plane-by-state} explains the state-level effects through the two RAS components ($\Delta\mathcal S$ and $\Delta\mathcal C$). The pooled state densities show a clear directional split: clean-to-wrong failures move toward worse confidence--correctness alignment, whereas wrong-to-correct recoveries move toward improved alignment. The highlighted model pair further shows why this component view is useful. Idefics3 and DeepSeek have similar response losses, but their state distributions differ: Idefics3 shows more polarized transition states, while DeepSeek shows weaker transition extremes with larger entropy shifts. Thus, similar response changes can arise from different structural patterns.

We then aggregate the state effects into source-level summaries for originally correct and originally wrong samples using Eq.~\eqref{eq:side-level-structural-effect}. The Mixed column in Table~\ref{tab:model-robustness-profile} shows that this tradeoff is common: many corruption--severity cells combine negative change on originally correct samples with positive change on originally wrong samples.

\begin{findingbox}{$2$}{Apparent accuracy gains are not always structural gains.}
Wrong-to-correct correction can coexist with degradation on originally correct samples, so positive response shifts need source attribution before being treated as robustness gains.
\end{findingbox}

The component view explains why the state effects differ. On originally correct samples, clean-to-wrong failures combine answer loss with worse confidence--correctness alignment, while still-correct cases can lose support without changing the top-$1$ answer. On originally wrong samples, wrong-to-correct correction is structurally meaningful when alignment improves, but this effect is diluted by the larger persistent-wrong group. Two sign patterns are especially informative: overconfidence, where alignment worsens while entropy decreases, and hesitant improvement, where alignment improves while entropy increases. These patterns show why corruption effects cannot be reduced to answer flips.

\subsection{Trajectory View: Severity Reveals Persistence Patterns}
\label{sec:model-trajectory-pathologies}
The state view explains where structural change occurs, but it averages over individual severity levels. It does not show whether a failure appears only once, reverses, or persists as corruption becomes stronger. We therefore group the five severities for each model--question--corruption triple into ordered trajectories. This yields $209{,}703$ complete trajectories. The trajectory analysis separates two quantities: the top-$1$ correctness path across severity and the RAS-based structural support within that path.

\begin{figure*}[t]
    \centering
    \includegraphics[width=\textwidth]{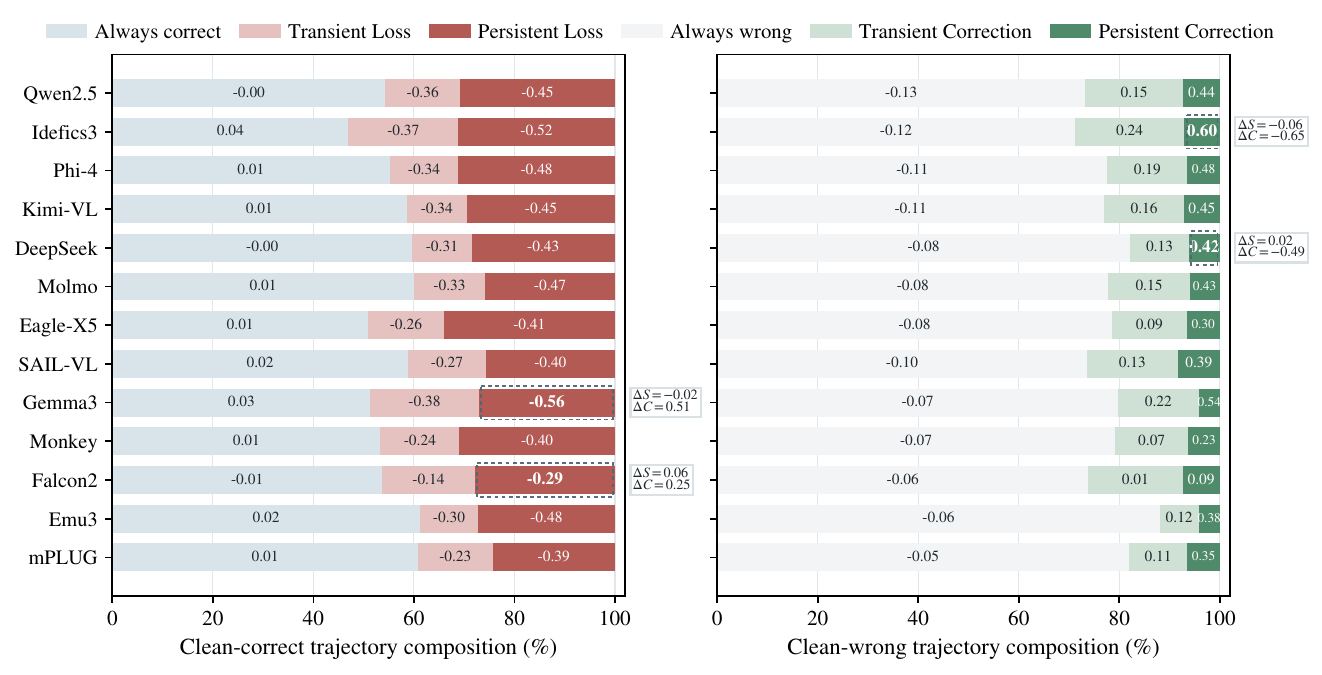}
    \caption{Model-level trajectory composition with class-conditioned RAS. Clean-correct trajectories use always correct, transient loss, and persistent loss labels; clean-wrong trajectories use always wrong, transient correction, and persistent correction labels. Segment numbers report mean class RAS; rows are ordered by mean trajectory RAS. Dashed outlines mark structurally different examples within the same class, with mean trajectory-median $\Delta\mathcal S$ and $\Delta\mathcal C$.}
    \label{fig:model-trajectory-composition}
\end{figure*}

For clean-correct trajectories, we distinguish \emph{always correct}, \emph{transient loss}, and \emph{persistent loss}; the latter two indicate whether a wrong prediction is temporary or persists after the first failure. For clean-wrong trajectories, we distinguish \emph{always wrong}, \emph{transient correction}, and \emph{persistent correction}; the correction labels indicate whether wrong-to-correct changes are temporary or remain correct through severity $5$. We also compute class-conditioned RAS for each group, turning answer-path labels into structural-support readings: persistent loss is more damaging than transient loss, and persistent correction is more supported than transient correction.

Fig.~\ref{fig:model-trajectory-composition} shows that aggregate degradation can arise from different trajectory compositions. Qwen2.5-VL has the strongest clean accuracy in Table~\ref{tab:model-robustness-profile}, but its trajectory profile reveals substantial clean-correct answer loss and negative class-conditioned RAS. Falcon2 has a smaller aggregate response loss, yet its low mixed-cell rate indicates a different source pattern rather than uniformly stronger structural robustness. The trajectory view separates answer loss, correction, and hidden degradation that aggregate metrics merge.

The dashed examples show the remaining gap between answer path and structural support. Within the same trajectory class, examples can have different $\Delta\mathcal S$, $\Delta\mathcal C$, and RAS values. On the clean-correct side, answer loss with decreased uncertainty and worse alignment indicates \emph{error overconfidence}, as illustrated by Gemma3 in Fig.~\ref{fig:model-trajectory-composition}. On the clean-wrong side, persistent correction can remain \emph{hesitant} when alignment improves but uncertainty also increases, as illustrated by DeepSeek. RAS-based attributes therefore mark silent degradation under preserved answers, overconfidence under answer loss, and weak support under correction.

\begin{findingbox}{$3$}{Same answer paths can differ in support.}
Answer paths can hide silent degradation, overconfident answer loss, or weakly supported correction; RAS attributes separate these cases within the same trajectory class.
\end{findingbox}

\subsection{Severity View: Condition and Model Effects}
\label{sec:corruption-severity}
Here, we ask whether stronger corruption only amplifies damage or also changes the balance between originally correct and wrong samples. We examine severity-conditioned trends across corruption families and models.

\begin{figure}[t]
    \centering
    \includegraphics[width=\columnwidth]{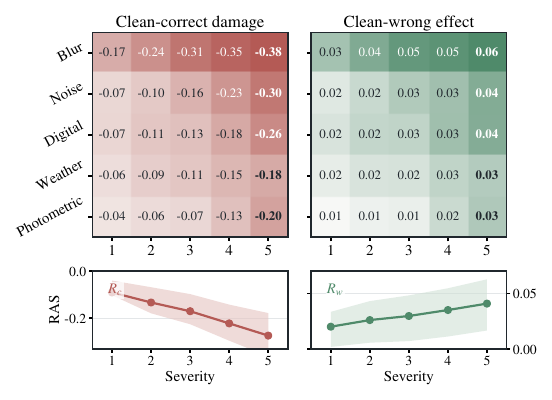}
    \caption{Condition-level diagnosis of severity effects. The matrices report signed clean-side RAS by corruption family and severity. Bottom insets show interquartile ranges across model--corruption cells: $R_c$ grows more negative with severity, while $R_w$ becomes more positive but remains smaller.}
    \label{fig:condition-model-diagnosis}
\end{figure}

Fig.~\ref{fig:condition-model-diagnosis} shows how source-level structural effects vary by corruption family, severity, and model. The clean-correct side is consistently negative and most damaging under blur, indicating that fine detail and contour loss strongly affect option-level support. The clean-wrong side is positive on average, but smaller than the clean-correct cost. Thus, the same corruption condition can damage originally correct predictions and weakly improve originally wrong ones.

The severity trends reveal the most counterintuitive pattern. As severity increases, the originally correct side becomes more negative, but the originally wrong side becomes more positive. Stronger corruption therefore appears to help some clean-wrong samples while causing larger clean-correct damage. We interpret this correction as prediction-structure fluctuation rather than robust correction: severe corruption can move some originally wrong predictions toward the correct answer while destabilizing originally correct predictions. This coupling is strongest for corruptions that disrupt spatial contours and fine visual detail.

\begin{findingbox}{$4$}{Stronger corruption can make wrong answers look better.}
Clean-wrong structural gains increase with severity, but they are coupled with larger clean-correct damage; this pattern is better interpreted as prediction-structure instability than robust correction.
\end{findingbox}


\subsection{Diagnostic Summary}
\label{sec:diagnostic-synthesis}
Overall, the results point to redistributed reliability under corruption. Aggregate metrics report the overall top-$1$ response change but hide two opposing effects: reduced support on originally correct samples and unstable movement toward correction on some originally wrong samples. The coarse-to-fine diagnosis separates these effects, checks their persistence across severity, and identifies when apparent gains coexist with clean-correct damage.





\section{Discussion}

\textit{Robustness failure is not only answer failure.}
Visual corruption can damage reliability before it changes the answer. A model may keep the correct top-$1$ option while losing support, or correct a wrong answer through unstable shifts. Aggregate scores can merge clean-correct cost with clean-wrong correction and mistake transient correction for robustness. Source, state, and trajectory attribution expose these cases.

\textit{Robustness training should preserve support, not only labels.}
These findings suggest that augmentation training should not reward only final-answer invariance. For originally correct samples, objectives should preserve correctness and confidence--correctness alignment. For originally wrong samples, reward correction mainly when the option distribution gains reliable support. Severity can supervise the four trajectory labels: transient loss, persistent loss, transient correction, and persistent correction.

\textit{Fine-detail evidence may be a bottleneck.}
The condition-level results suggest a design hypothesis. Blur causes clean-correct structural damage, consistent with failures to preserve contours, small text, and local grounding cues. This does not prove an internal mechanism, but suggests that robustness may require stable fine-grained evidence under degradation.





\section{Limitations}

\textit{Scope.}
\textsc{Bench-C} is a corruption-responsive testbed, not an estimate of natural failure prevalence. This scope makes structural changes easier to compare under matched clean--corrupted inputs, but the measured patterns should be read as diagnostic failure modes rather than open-world frequencies. Future work can apply the same analysis to broader or naturally sampled data.

\textit{Corruption coverage.}
The testbed uses controlled single-factor corruptions with ordered severities. This supports paired diagnosis, but does not cover mixed in-the-wild distortions, camera artifacts, compression chains, or other acquisition effects. Extending the protocol would test whether the same patterns hold under realistic degradation.

\textit{Closed-source models.}
The main analysis uses option-level logits, which are not always available for closed-source VLMs. A black-box variant can estimate option distributions through repeated sampling, and our code supports this route. This extends the diagnosis beyond open models, but adds cost and requires checks on sample count, temperature, and prompts.

\section{Conclusion}

This paper presents \textsc{Bench-C}, a compact diagnostic testbed for corruption-induced reliability failures in VLMs. It identifies samples where corruption exposes model differences, measures paired changes in the option distribution, and attributes those changes by source, state, and severity trajectory. Experiments across $13$ VLMs, $19$ corruptions, and five severity levels show that corruption effects are not a single degradation trend: preserved answers can lose support, apparent accuracy gains can coexist with clean-correct damage, and the same answer path can carry different structural evidence. Evaluating both response and distribution levels shows where damage, correction, and structural support occur under corruption.

\bibliographystyle{IEEEtran}
\bibliography{main}

\ifarxiv
\clearpage
\markboth{Submitted to IEEE Transactions on Multimedia}%
{Diagnosing Corruption-Induced Reliability Failures in Vision-Language Models}
\setcounter{section}{0}
\setcounter{figure}{7}
\setcounter{table}{3}
\section*{Supplementary Material}
\section{Supplement Overview}

We further validate three aspects of our study:
\begin{itemize}
    \item \textsc{Bench-C} preserves broad source and semantic coverage while enriching corruption-responsive samples;
    \item RAS exposes reliability changes hidden by response-level metrics;
    \item Model-level robustness gaps can be traced from response metrics to structural trajectories.
\end{itemize}

\section{Bench-C Testbed and Selection Reliability}

We check that the selection is both broad and diagnostic: broad across sources and semantic categories, and diagnostic in the sense that it emphasizes corruption-responsive samples.

\subsection{Coverage as Diagnostic Density}
Fig.~\ref{fig:supp-benchc-composition} checks whether \textsc{Bench-C} still covers multiple data sources and semantic types after selecting corruption-responsive samples.
As shown in Fig.~\ref{fig:supp-benchc-composition}(a), the split retains all six source benchmarks, reducing the risk that later robustness conclusions are driven by a single source.
The source shares also reflect diagnostic density: \textsc{SeedBench2+} contributes the largest share because its broad task mix contains many corruption-sensitive visual cues, whereas \textsc{LEGO-Puzzles} is visually concentrated on LEGO-style objects and mainly geometric-rotation questions, so the semantic-diversity constraint naturally limits its share.
At the same time, \textsc{Bench-C} is not intended to preserve every category in proportion to the candidate pool.
Fig.~\ref{fig:supp-benchc-composition}(b) shows the same logic at the L2-category level: categories such as web, image retrieval, map, chart, and complex OCR are more useful for corruption diagnosis because they depend on fine detail, layout, text-like evidence, or candidate comparison.
Lower-count categories are not invalid, but many are more specialized, sparse, or less directly tied to the corruption types used here.
Thus, \textsc{Bench-C} remains broad enough to avoid a narrow benchmark artifact while being reweighted toward samples where visual degradation can expose reliability changes.

\begin{figure}[t]
    \centering
    \includegraphics[width=0.90\columnwidth]{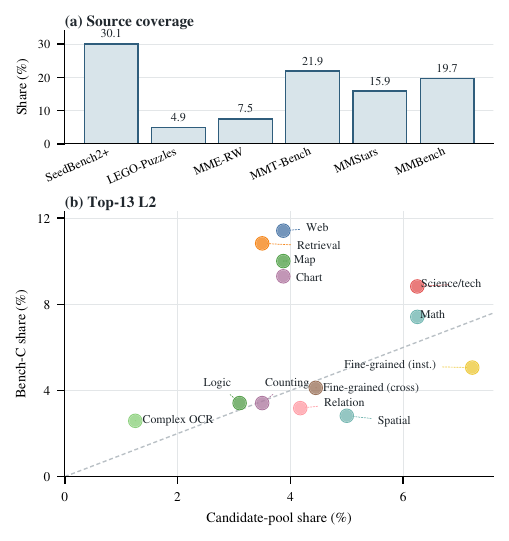}
    \caption{\textsc{Bench-C} composition. (a) The selected split keeps all six source benchmarks, reducing single-source bias. (b) Thirteen L2 categories are shown: the top-12 by \textsc{Bench-C} share plus Complex OCR. Colored points distinguish categories, with dashed leaders for dense labels. The split stays broad while concentrating more mass on categories where corruptions affect fine detail, layout, text-like evidence, or candidate comparison.}
    \label{fig:supp-benchc-composition}
\end{figure}

\subsection{Responsiveness of the Selection Signal}

The main paper shows that selector models can reweight the candidate pool toward more informative samples.
Here we ask whether this property remains visible in a broader evaluation setting.
Fig.~\ref{fig:supp-benchc-examples}(a) compares the construction-time $\kappa(q)$ distribution with the same top-$1$ answer Gini computed from the $13$ tested models on \textsc{Bench-C}.
The selected split shifts toward higher $\kappa(q)$ under the selector models, and the tested-model distribution occupies the same high-response region rather than collapsing back to the low-inconsistency candidate-pool mass.
This agreement suggests that selector-model inconsistency is a usable selection signal: the selected samples remain corruption-responsive under a larger VLM set, instead of matching only the selectors.

\begin{figure}[t]
    \centering
    {\footnotesize\textbf{(a)} $\kappa(q)$ distributions\par}
    \includegraphics[width=0.92\columnwidth]{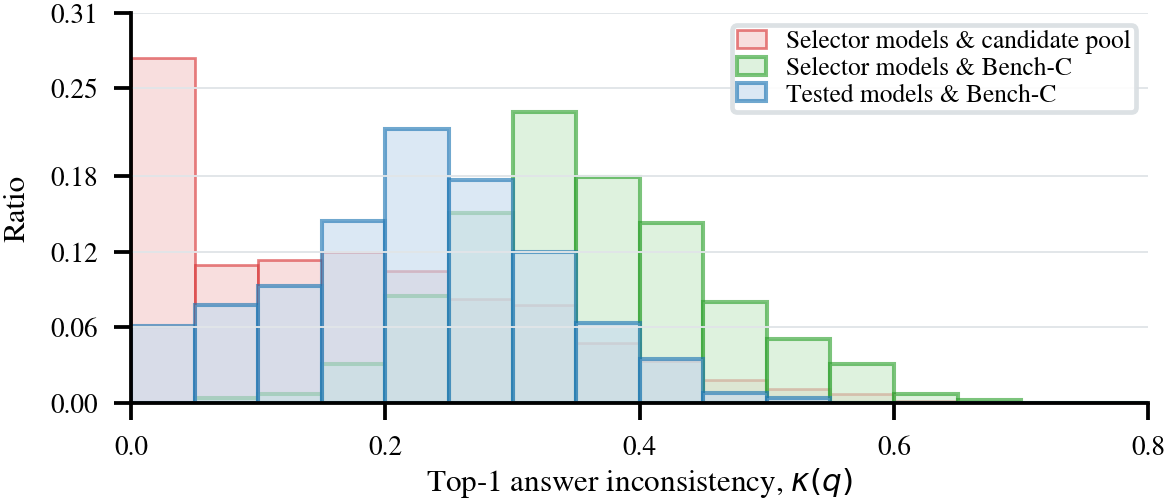}
    \vspace{1pt}
    \makebox[\columnwidth][c]{%
    \begin{minipage}[t]{0.325\columnwidth}
        \centering
        {\scriptsize\textbf{(b)} low $\kappa(q)$, $\kappa=0.066$\par \vspace{2pt}}
        \includegraphics[width=\linewidth]{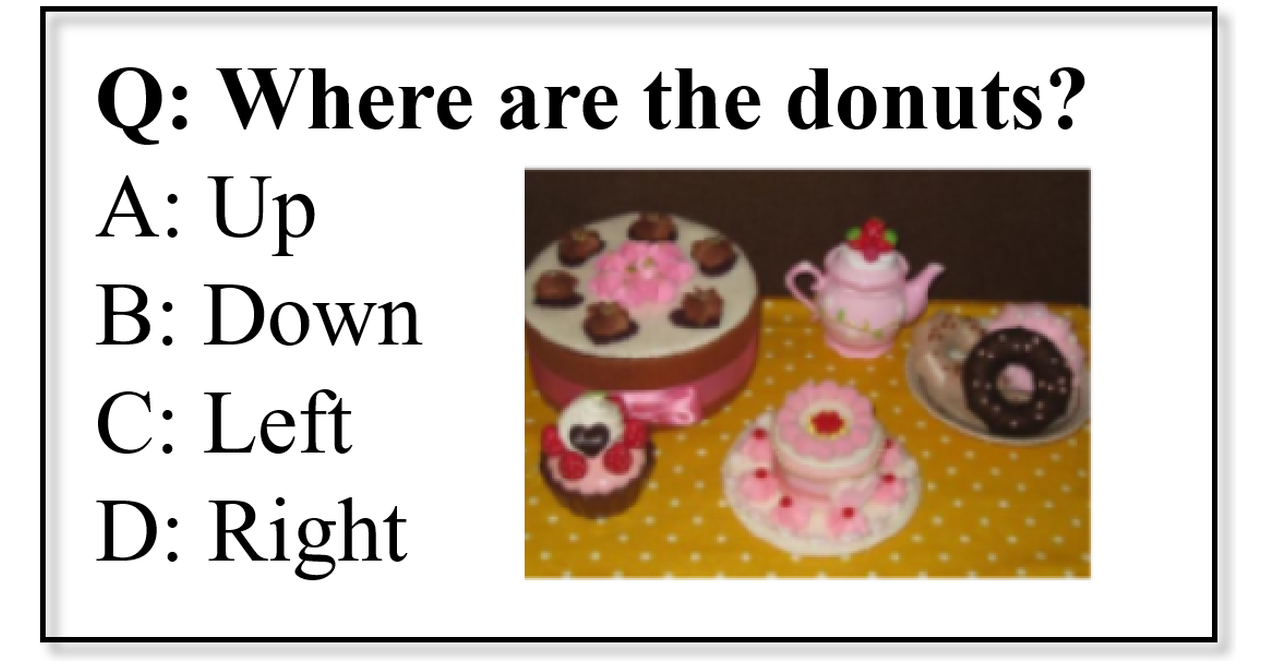}
    \end{minipage}%
    \hspace{0.055\columnwidth}%
    \begin{minipage}[t]{0.365\columnwidth}
        \centering
        {\scriptsize\textbf{(c)} medium $\kappa(q)$, $\kappa=0.356$\par}
        \includegraphics[width=\linewidth]{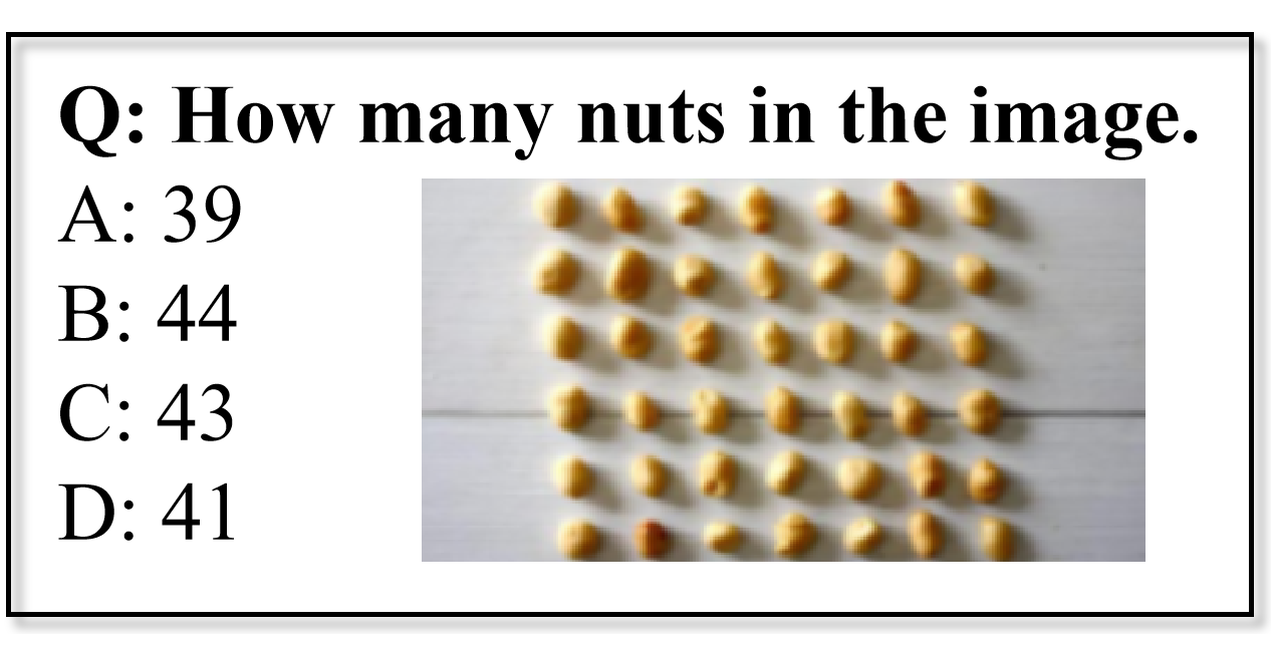}
    \end{minipage}%
    }
    \\ \vspace{1pt}
    {\scriptsize\textbf{(d)} high $\kappa(q)$, $\kappa=0.589$\par \vspace{2pt}}
    \includegraphics[width=0.76\columnwidth]{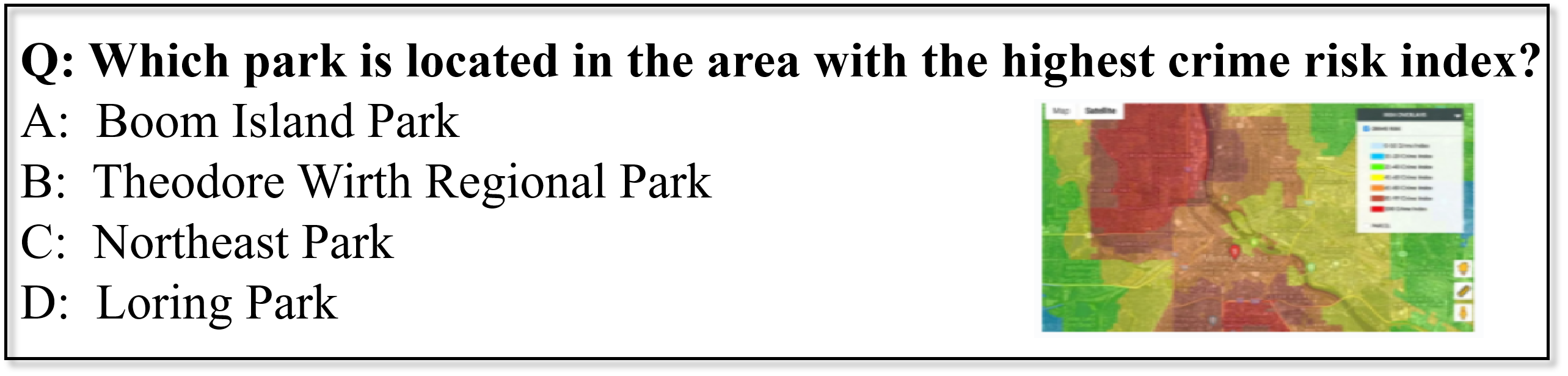}
    \caption{Selection-signal reliability. (a) The construction-time $\kappa(q)$ distribution for the candidate pool and \textsc{Bench-C} is compared with the same top-$1$ answer Gini computed from the $13$ tested models on \textsc{Bench-C}. (b)--(d) Candidate examples at low, medium, and high $\kappa(q)$ show that larger inconsistency tends to correspond to more quality-sensitive visual evidence.}
    \label{fig:supp-benchc-examples}
\end{figure}

The case examples give the distributional shift a semantic reading.
The low-$\kappa(q)$ example in Fig.~\ref{fig:supp-benchc-examples}(b) is visually simpler, whereas the medium- and high-$\kappa(q)$ examples in Fig.~\ref{fig:supp-benchc-examples}(c)--(d) require counting quality and map-detail evidence.
In such cases, controlled corruption can change the top-$1$ answer and its distributional support.

\section{RAS Component Diagnostics}
RAS exposes prediction-structure changes hidden by answer labels and response metrics. The following analyses check whether the two components produce recurring regimes and how those regimes change the reading of answer paths.

\subsection{Component Regimes and Case-Level Reading}

The RAS components induce four regimes from confidence--correctness alignment and uncertainty changes.
We assign $1{,}048{,}478$ valid model--corruption--severity--sample pairs to these regimes by the signs of $\Delta\mathcal C$ and $\Delta\mathcal S$.
Table~\ref{tab:supp-ras-validation} shows that all four non-boundary regimes have substantial support, indicating that the component plane provides recurring diagnostic regions rather than sparse exceptions.
The signs should be read jointly rather than as globally favorable directions: $\Delta\mathcal C$ identifies whether alignment improves or worsens, while $\Delta\mathcal S$ identifies whether the distribution becomes flatter or sharper.
Interpreted together with the answer state, these directions separate reliability readings such as overconfident degradation when worsening alignment is accompanied by sharpening, and hesitant support when improved alignment is accompanied by flattening.
Using either component alone would merge these distinctions; RAS preserves them through a joint alignment--uncertainty characterization.

\begin{table}[t]
\centering
\caption{Empirical RAS sign regimes over $1{,}048{,}478$ valid clean--corrupted pairs. Rows are grouped by the signs of $\Delta\mathcal C$ and $\Delta\mathcal S$; each row reports regime means.}
\label{tab:supp-ras-validation}
\scriptsize
\setlength{\tabcolsep}{4pt}
\renewcommand{\arraystretch}{1.08}
\begin{tabular}{@{}lrrrr@{}}
\toprule
Regime & $\overline{\Delta\mathcal C}$ & $\overline{\Delta\mathcal S}$ & $\overline{\mathrm{RAS}}$ & Share \\
\midrule
$\Delta\mathcal C>0$ worse, $\Delta\mathcal S>0$ flatter & $+0.258$ & $+0.138$ & $-0.258$ & $28.36\%$ \\
$\Delta\mathcal C>0$ worse, $\Delta\mathcal S<0$ sharper & $+0.188$ & $-0.124$ & $-0.227$ & $22.58\%$ \\
$\Delta\mathcal C<0$ better, $\Delta\mathcal S>0$ flatter & $-0.150$ & $+0.134$ & $+0.112$ & $23.71\%$ \\
$\Delta\mathcal C<0$ better, $\Delta\mathcal S<0$ sharper & $-0.144$ & $-0.095$ & $+0.144$ & $13.39\%$ \\
Boundary & $+0.009$ & $+0.000$ & $-0.009$ & $11.96\%$ \\
\bottomrule
\end{tabular}
\end{table}

\begin{figure}[t]
    \centering
    \includegraphics[width=0.82\columnwidth]{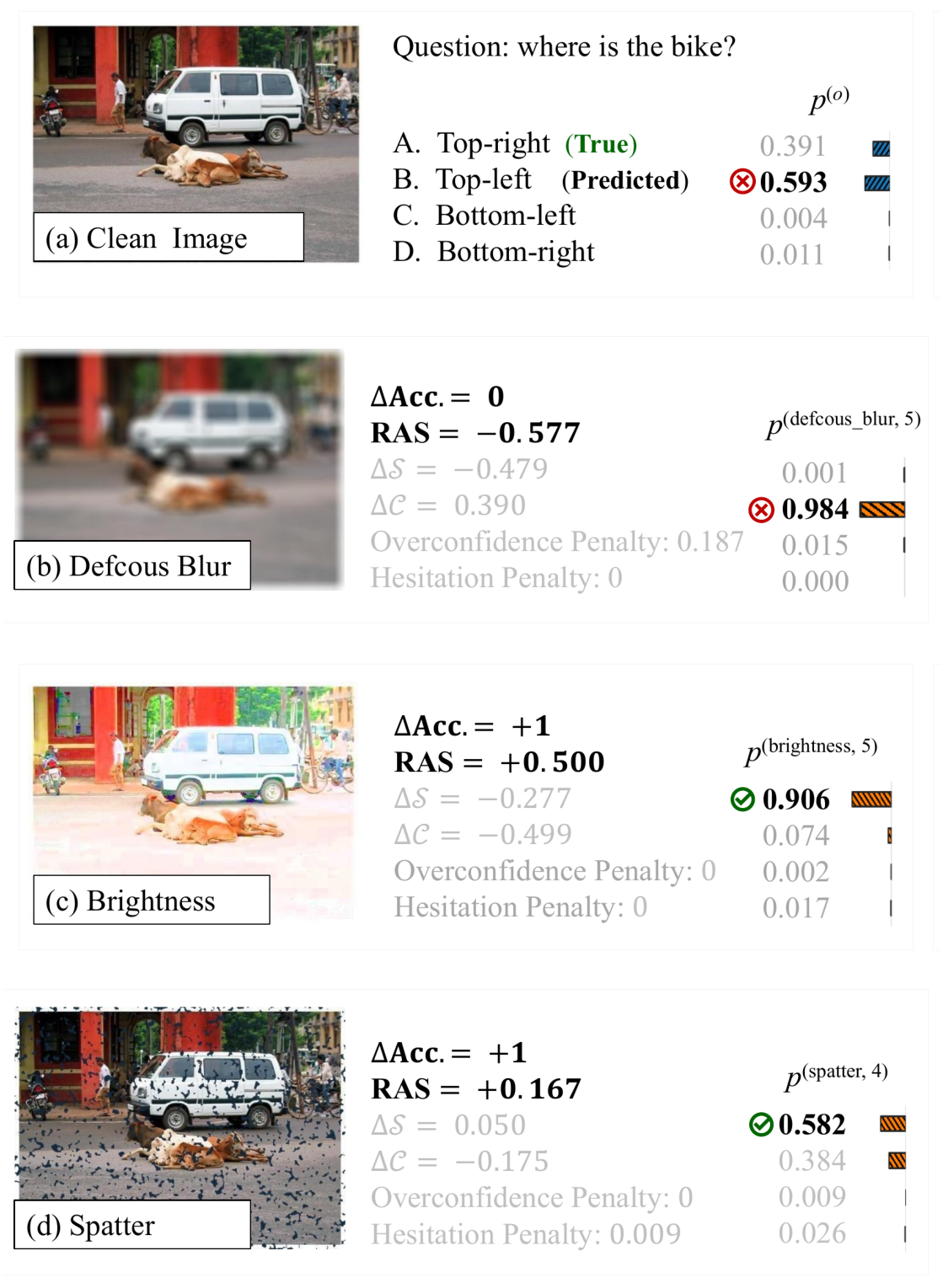}
    \caption{Case-level RAS behavior of DeepSeek-VL2-Small. From the same clean question, defocus blur keeps the answer wrong but overconfident, while brightness and spatter correct it with different structural support. RAS captures these option-distribution changes rather than treating all top-$1$ transitions alike.}
    \label{fig:supp-ras-example}
\end{figure}

Fig.~\ref{fig:supp-ras-example} gives a case-level example of why RAS differs from response metrics.
For the same model and clean question, response metrics can only record whether the top-$1$ answer stays wrong or becomes correct.
RAS additionally describes how the option distribution supports that answer under each corruption.
Starting from the same clean question in Fig.~\ref{fig:supp-ras-example}(a), defocus blur keeps the answer wrong in Fig.~\ref{fig:supp-ras-example}(b), whereas brightness and spatter recover the correct answer in Fig.~\ref{fig:supp-ras-example}(c)--(d).
The component values add a reliability reading to these answer paths: a wrong answer is more concerning when confidence concentrates on it, and recovered answers can differ in how much distributional support they gain.
The component regimes also summarize model behavior: under the same top-$1$ transition, Fig.~\ref{fig:supp-model-component-signature} shows models occupying different component regions, making the rank mismatches in Table~\ref{tab:supp-model-profile} diagnostic rather than merely numerical.

\subsection{Closed-Source RAS}

RAS can also be applied when option logits are not directly available.
For a black-box API without logits, we query the same multiple-choice prompt repeatedly with temperature $>0$, map each returned answer to an option, and use the empirical option frequencies as the diagnostic distribution.
Table~\ref{tab:supp-blackbox-ras} shows that this sampling proxy closely matches direct-logit RAS on Qwen3-VL-Plus, supporting black-box use of the same RAS components.

\begin{table}[t]
\centering
\refstepcounter{table}
\label{tab:supp-blackbox-ras}
{\footnotesize TABLE~\thetable\\[-1pt]
\textsc{Closed-source RAS diagnostics. Sampling estimates option distributions from repeated stochastic decoding.}\par}
\vspace{2pt}
\scriptsize
\setlength{\tabcolsep}{3.4pt}
\renewcommand{\arraystretch}{1.05}
\begin{tabular}{@{}lcccc@{}}
\toprule
Model & Samples & Acc.$^{(o)}\uparrow$ & $\Delta$Acc$\uparrow$ & RAS$\uparrow$ \\
\midrule
ChatGPT-4 Omni & $100{\times}19{\times}5$ & $0.680$ & $-0.076$ & $-0.106$ \\
Qwen3-VL-Plus & $100{\times}19{\times}5$ & $0.679$ & $-0.084$ & $-0.117$ \\
Qwen3-VL-Plus-Logit & $100{\times}1{\times}5$ & $0.690$ & $-0.022$ & $-0.037$ \\
Qwen3-VL-Plus-Sampling & $100{\times}1{\times}5$ & $0.680$ & $-0.027$ & $-0.036$ \\
\bottomrule
\end{tabular}
\end{table}

\section{Model-Level Reliability Analysis}

This section uses the RAS to examine model-level reliability gaps. 
The goal is to identify when response-based and structural robustness give different readings of the same models.

\subsection{Rank Mismatch Across Metrics}

The model-level analysis uses the RAS evidence above to ask where response-based rankings and structural rankings disagree across models.
Table~\ref{tab:supp-model-profile} reports each response metric as score with superscript rank and rank shift to RAS; the RAS column reports score and rank. The rows follow the RAS ranking, so the table first shows the structural ordering and then marks where response metrics depart from it.

\begin{table}[!t]
\centering
\refstepcounter{table}
\label{tab:supp-model-profile}
{\footnotesize
\textsc{TABLE~\thetable. Model-level rank mismatch; RAS-ordered rows.}\\[-1pt]
\textsc{Non-RAS cells add $\Delta r=r_{\mathrm{RAS}}-r_m$ (rank $1$ best).}\par}
\vspace{2pt}
\scriptsize
\setlength{\tabcolsep}{2.2pt}
\renewcommand{\arraystretch}{1.2}
\begin{tabular*}{\linewidth}{@{\extracolsep{\fill}}lcccc@{}}
\toprule
Model & Abs.$\uparrow$ & $\Delta$Acc$\uparrow$ & Rel.$\uparrow$ & RAS$\uparrow$ \\
\midrule
mPLUG & $0.416^{8,\textcolor{blue!70!black}{\mathbf{-7}}}$ & $-0.049^{3,\textcolor{blue!70!black}{\mathbf{-2}}}$ & $0.894^{2,\textcolor{blue!70!black}{\mathbf{-1}}}$ & $-0.049^{1}$ \\
Emu3 & $0.311^{13,\textcolor{blue!70!black}{\mathbf{-11}}}$ & $-0.041^{2,+0}$ & $0.884^{4,\textcolor{blue!70!black}{\mathbf{-2}}}$ & $-0.057^{2}$ \\
Falcon2 & $0.400^{9,\textcolor{blue!70!black}{\mathbf{-6}}}$ & $-0.040^{1,\textcolor{red!70!black}{\mathbf{+2}}}$ & $0.909^{1,\textcolor{red!70!black}{\mathbf{+2}}}$ & $-0.058^{3}$ \\
Monkey & $0.365^{12,\textcolor{blue!70!black}{\mathbf{-8}}}$ & $-0.052^{4,+0}$ & $0.875^{5,\textcolor{blue!70!black}{\mathbf{-1}}}$ & $-0.068^{4}$ \\
Gemma3 & $0.367^{11,\textcolor{blue!70!black}{\mathbf{-6}}}$ & $-0.061^{5,+0}$ & $0.858^{7,\textcolor{blue!70!black}{\mathbf{-2}}}$ & $-0.073^{5}$ \\
SAIL-VL & $0.517^{3,\textcolor{red!70!black}{\mathbf{+3}}}$ & $-0.063^{6,+0}$ & $0.892^{3,\textcolor{red!70!black}{\mathbf{+3}}}$ & $-0.073^{6}$ \\
Eagle-X5 & $0.383^{10,\textcolor{blue!70!black}{\mathbf{-3}}}$ & $-0.069^{7,+0}$ & $0.848^{9,\textcolor{blue!70!black}{\mathbf{-2}}}$ & $-0.075^{7}$ \\
Molmo & $0.494^{6,\textcolor{red!70!black}{\mathbf{+2}}}$ & $-0.072^{8,+0}$ & $0.873^{6,\textcolor{red!70!black}{\mathbf{+2}}}$ & $-0.079^{8}$ \\
DeepSeek & $0.504^{5,\textcolor{red!70!black}{\mathbf{+4}}}$ & $-0.089^{9,+0}$ & $0.850^{8,\textcolor{red!70!black}{\mathbf{+1}}}$ & $-0.088^{9}$ \\
Kimi-VL & $0.534^{2,\textcolor{red!70!black}{\mathbf{+8}}}$ & $-0.096^{11,\textcolor{blue!70!black}{\mathbf{-1}}}$ & $0.848^{10,+0}$ & $-0.098^{10}$ \\
Phi-4 & $0.506^{4,\textcolor{red!70!black}{\mathbf{+7}}}$ & $-0.098^{12,\textcolor{blue!70!black}{\mathbf{-1}}}$ & $0.837^{11,+0}$ & $-0.101^{11}$ \\
Idefics3 & $0.466^{7,\textcolor{red!70!black}{\mathbf{+5}}}$ & $-0.094^{10,\textcolor{red!70!black}{\mathbf{+2}}}$ & $0.832^{12,+0}$ & $-0.101^{12}$ \\
Qwen2.5 & $0.552^{1,\textcolor{red!70!black}{\mathbf{+12}}}$ & $-0.115^{13,+0}$ & $0.828^{13,+0}$ & $-0.118^{13}$ \\
\bottomrule
\end{tabular*}
\end{table}

Table~\ref{tab:supp-model-profile} exposes two disagreement patterns.
Red cells mark response views that rank a model higher than RAS does, indicating that the corresponding response metric overestimates structural robustness for that model.
Qwen2.5 is ranked first by absolute robustness (corrupted accuracy) but last by RAS, and Kimi-VL and Phi-4 show similar demotion under absolute robustness.
These models still answer many corrupted questions correctly, but their matched clean--corrupted distributions lose more support.
This is silent structural damage at the model level: the top-$1$ answer often survives while the distribution behind it becomes less reliable.

Blue cells show the complement: the response metric ranks the model lower than RAS does.
Emu3, Monkey, and mPLUG are promoted by RAS in several views because their corruption-induced changes are less structurally destructive.
This does not mean that they are more capable; rather, capability and corruption response are separable.
A model can answer fewer questions correctly while preserving the structure of the answers it keeps.
Falcon2 gives a useful middle case.
It has the best $\Delta$Acc and relative-robustness ranks, but only the third-best RAS.
Answer preservation and support preservation are therefore related, but not interchangeable.
The rank shifts in Table~\ref{tab:supp-model-profile} should be read as diagnostic disagreements rather than as a replacement leaderboard.

\begin{figure}[!t]
    \centering
    \includegraphics[width=\columnwidth]{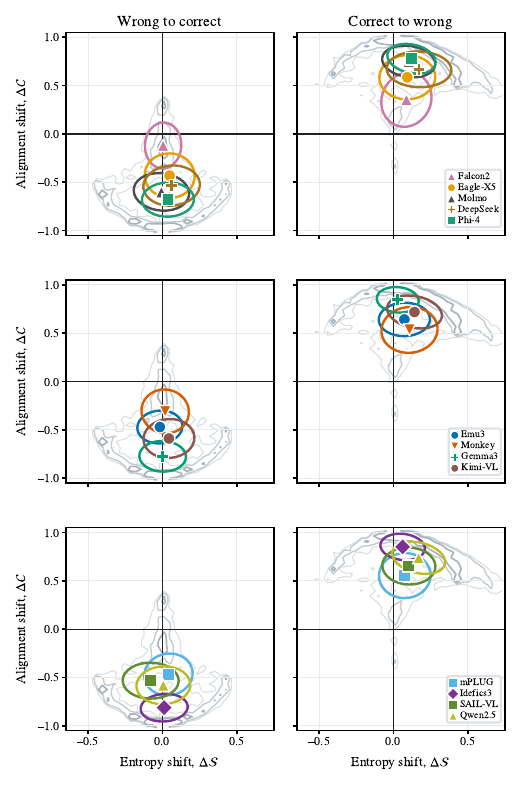}
    \caption{RAS component signatures separate model-level failure patterns. Rows partition the $13$ models into signature-diverse sets, and columns show wrong-to-correct and correct-to-wrong transitions in the $(\Delta\mathcal S,\Delta\mathcal C)$ plane. Gray contours show the pooled distribution of all model--sample--corruption pairs in the corresponding transition state; colored ellipses summarize the core distribution of each model.}
    \label{fig:supp-model-component-signature}
\end{figure}

\subsection{Component Signatures Behind Model Gaps}

This subsection asks whether a fixed top-$1$ transition still hides model differences.
In Fig.~\ref{fig:supp-model-component-signature}, each column fixes the visible transition: wrong-to-correct on the left and correct-to-wrong on the right.
Yet the model ellipses occupy different regions of the $(\Delta\mathcal S,\Delta\mathcal C)$ plane.
Correction can therefore be supported or hesitant, and answer loss can differ in whether uncertainty or alignment is more affected.
This component view explains the rank mismatches in Table~\ref{tab:supp-model-profile}.
RAS-promoted models avoid destructive component shifts despite lower absolute accuracy, whereas RAS-demoted models preserve many answers while moving toward less reliable signatures.
Falcon2 and mPLUG provide a compact example.
Table~\ref{tab:supp-model-profile} ranks Falcon2 first by $\Delta$Acc but third by RAS, while mPLUG ranks third by $\Delta$Acc but first by RAS.
Fig.~\ref{fig:supp-model-component-signature} explains this reversal from the component side.
Falcon2's advantage is mainly response-level: its accuracy loss is smaller, as reflected by the best $\Delta$Acc rank.
Within the fixed transition states, however, mPLUG shows more favorable component signatures, especially on the correction side where wrong-to-correct cases remain better supported.
Thus, $\Delta$Acc counts answer outcomes, whereas RAS also reflects the uncertainty--alignment structure behind those outcomes.
Qwen2.5 and mPLUG show the same gap from the absolute-robustness side.
Qwen2.5 ranks first in corrupted accuracy but last by RAS; its correction-side signature can be favorable, but its answer-loss signature shifts more along $\Delta\mathcal S>0$.
Relative to Qwen2.5, mPLUG stays closer to the origin across the two transition types, indicating smaller changes in uncertainty and alignment in this comparison.
These comparisons show that model gaps are not determined only by how often top-$1$ states change; they also depend on the component signatures carried by the same visible transitions.

\subsection{Trajectory Sources of Model Gaps}

The previous subsection fixes one-step top-$1$ transitions.
We next ask whether the same severity-level trajectory can still hide different structural behavior.
Fig.~\ref{fig:supp-trajectory-components} visualizes two sources of model-level gaps: trajectory mass, shown by how often samples enter a trajectory class, and within-trajectory support, shown by the class-conditioned $(\Delta\mathcal S,\Delta\mathcal C)$ signature.
The trajectory labels are answer-path labels only: \emph{Transient Loss} and \emph{Persistent Loss} describe clean-correct samples that become wrong at some severity, whereas \emph{Transient Correction} and \emph{Persistent Correction} describe clean-wrong samples that become correct at some severity.

\begin{figure*}[t]
    \centering
    \includegraphics[width=\textwidth]{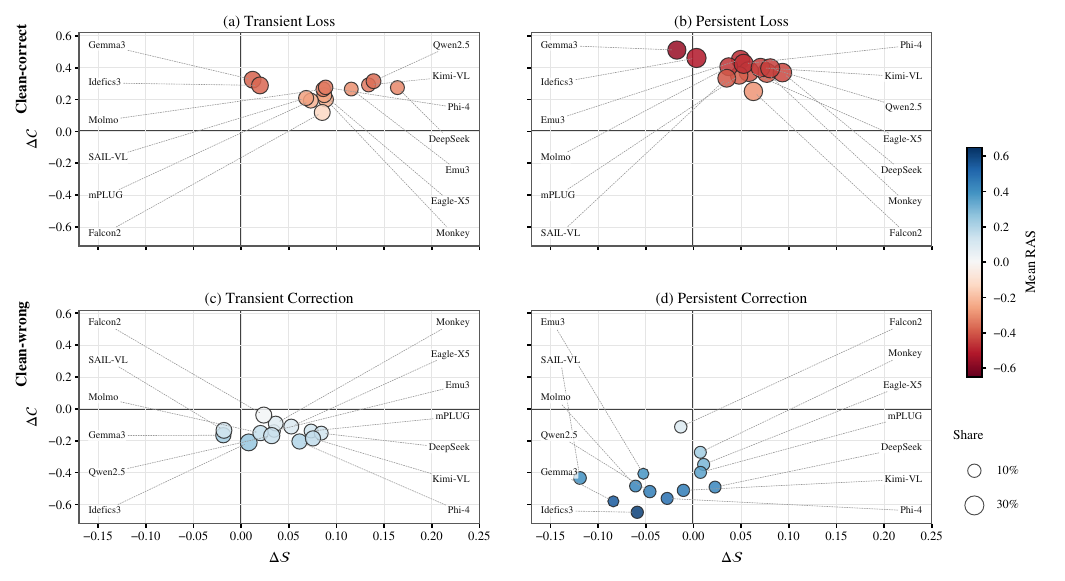}
    \caption{Trajectory-level component signatures reveal model gaps within the same visible answer path. Each panel fixes one trajectory class and plots model-level means in the $(\Delta\mathcal S,\Delta\mathcal C)$ plane. Marker size indicates trajectory share, while color and position indicate structural support carried by that trajectory.}
    \label{fig:supp-trajectory-components}
\end{figure*}

On the clean-correct side, answer-loss trajectories are not structurally neutral.
For \emph{Transient Loss}, all models lie in the first quadrant, so even temporary answer loss is accompanied by flatter uncertainty and worse confidence--correctness alignment.
\emph{Persistent Loss} is more damaging and more model-specific: most models remain in the first quadrant, while Gemma3 moves into the second quadrant, where alignment worsens as uncertainty becomes sharper.
This is an overconfident answer-loss pattern, in which the wrong answer is supported by a more concentrated distribution rather than by appropriate caution.
Qwen2.5 also has a large \emph{Persistent Loss} share in a damaging region, explaining how high corrupted accuracy can coexist with weak structural robustness.

The clean-wrong trajectories show the correction side.
\emph{Transient Correction} improves alignment, but many models remain near or to the right of $\Delta\mathcal S=0$, indicating correction with limited uncertainty support.
\emph{Persistent Correction} is more structurally supported: most models move downward in $\Delta\mathcal C$, and several also move left in $\Delta\mathcal S$.
Falcon2 remains close to the origin, consistent with the earlier observation that its response-level advantage does not translate into the strongest correction-side support.
Thus, correction trajectories differ not only in how often they occur, but also in whether the corrected answer is structurally supported.

The figure should be read as a joint map.
Marker size identifies trajectory mass, the horizontal coordinate shows whether corruption flattens or sharpens the option distribution, and the vertical coordinate shows whether confidence--correctness alignment worsens or improves.
A large marker in a damaging region indicates frequent entry into a harmful answer path, whereas a smaller marker far from the origin indicates a rarer but structurally severe path.
Conversely, points near the origin can reflect visible answer-path changes whose component movement remains limited.

Reading Fig.~\ref{fig:supp-model-component-signature} with Fig.~\ref{fig:supp-trajectory-components} helps distinguish mass-related and support-related gaps.
In Fig.~\ref{fig:supp-trajectory-components}, a mass-related difference appears as marker-size differences within a trajectory class, whereas a support-related difference appears as different locations or colors among models with comparable trajectory mass.
The trajectory view adds a temporal source to the rank gap: structural weakness can come from entering harmful paths more often, from weaker support within the same path, or from both.
Read with Table~\ref{tab:supp-model-profile} and Fig.~\ref{fig:supp-model-component-signature}, it separates models with similar scalar ranks but different failure profiles.

\fi

\end{document}